\documentclass[10pt,twocolumn,letterpaper]{article}

\usepackage{wacv}
\wacvfinalcopy %
\usepackage{times}
\usepackage{epsfig}
\usepackage{graphicx}
\usepackage{amsmath}
\usepackage{amssymb}
\usepackage{multirow}
\usepackage{xcolor}
\usepackage{subcaption}
\usepackage{array}

\usepackage[export]{adjustbox}

\newcolumntype{P}[1]{>{\centering\arraybackslash}p{#1}}

\def\wacvPaperID{0033} %

\ifwacvfinal
\def\assignedStartPage{1} %
\fi

\ifwacvfinal
\usepackage[breaklinks=true,bookmarks=false]{hyperref}
\else
\usepackage[pagebackref=true,breaklinks=true,colorlinks,bookmarks=false]{hyperref}
\fi

\ifwacvfinal
\else
\fi

\begin{document}

\title{DeepOpht: Medical Report Generation for Retinal Images via Deep Models and Visual Explanation}

\author{Jia-Hong Huang$^{1\dagger	}$,~C.-H. Huck Yang$^{2\ddagger	,3}$,~Fangyu Liu$^{4}$,~Meng Tian$^{5}$,~Yi-Chieh Liu$^{3,9}$,~Ting-Wei Wu$^{3,7}$, \\I-Hung Lin$^{6}$,~Kang Wang$^{8}$,~Hiromasa Morikawa$^{2}$,~Hernghua Chang$^{9}$,~Jesper Tegner$^{2}$,~Marcel Worring$^{1}$ \\
\small$^{1}$University of Amsterdam, $^{2}$King Abdullah University of Science and Technology (KAUST), $^{3}$Georgia Institute of Technology, \\ \small $^{4}$University of Cambridge, $^{5}$Department of Ophthalmology, Bern University Hospital, $^{6}$Tri-Service General Hospital,\\ 
\small $^{7}$University of California, Berkeley, $^{8}$Beijing Friendship Hospital, and $^{9}$National Taiwan University \\
{\tt\small $\dagger$ corresponding author: j.huang@uva.nl $\ddagger$ work conducted during visiting KAUST.}
}

\maketitle

\begin{abstract}
    In this work, we propose an AI-based method that intends to improve the conventional retinal disease treatment procedure and help ophthalmologists increase diagnosis efficiency and accuracy. The proposed method is composed of a deep neural networks-based (DNN-based) module, including a retinal disease identifier and clinical description generator, and a DNN visual explanation module.
    To train and validate the effectiveness of our DNN-based module, we propose a large-scale retinal disease image dataset. Also, as ground truth, we provide a retinal image dataset manually labeled by ophthalmologists to qualitatively show, the proposed AI-based method is effective. With our experimental results, we show that the proposed method is quantitatively and qualitatively effective. Our method is capable of creating meaningful retinal image descriptions and visual explanations that are clinically relevant. 
    \textbf{\href{https://github.com/Jhhuangkay/DeepOpht-Medical-Report-Generation-for-Retinal-Images-via-Deep-Models-and-Visual-Explanation}{DeepOpht Github.}}
\end{abstract}

\section{Introduction}
The World Health Organization (WHO) estimates that typical retinal diseases such as Age-related Macular Degeneration (AMD) and Diabetic Retinopathy (DR) are expected to affect over 500 million people worldwide shortly \cite{pizzarello2004vision}. Besides, generally speaking, the traditional process of retinal disease diagnosis and creating a medical report for a patient takes time in practice. The above means that ophthalmologists will become busier and busier. 

As we may know, the current state of the art in Artificial Intelligence (AI) involves deep learning research, and we claim deep learning is one of the promising ways to help ophthalmologists and improve the traditional retinal disease treatment procedure. Deep learning based models such as convolutional neural networks (CNN) or recurrent neural networks (RNN) for computer vision or natural language processing tasks, respectively, have achieved, and, in some cases, even exceeded human-level performance. There is no better time than now to propose an AI-based medical diagnosis method to aid ophthalmologists.

\begin{figure}[t!]
\begin{center}
\includegraphics[width=1.0\linewidth]{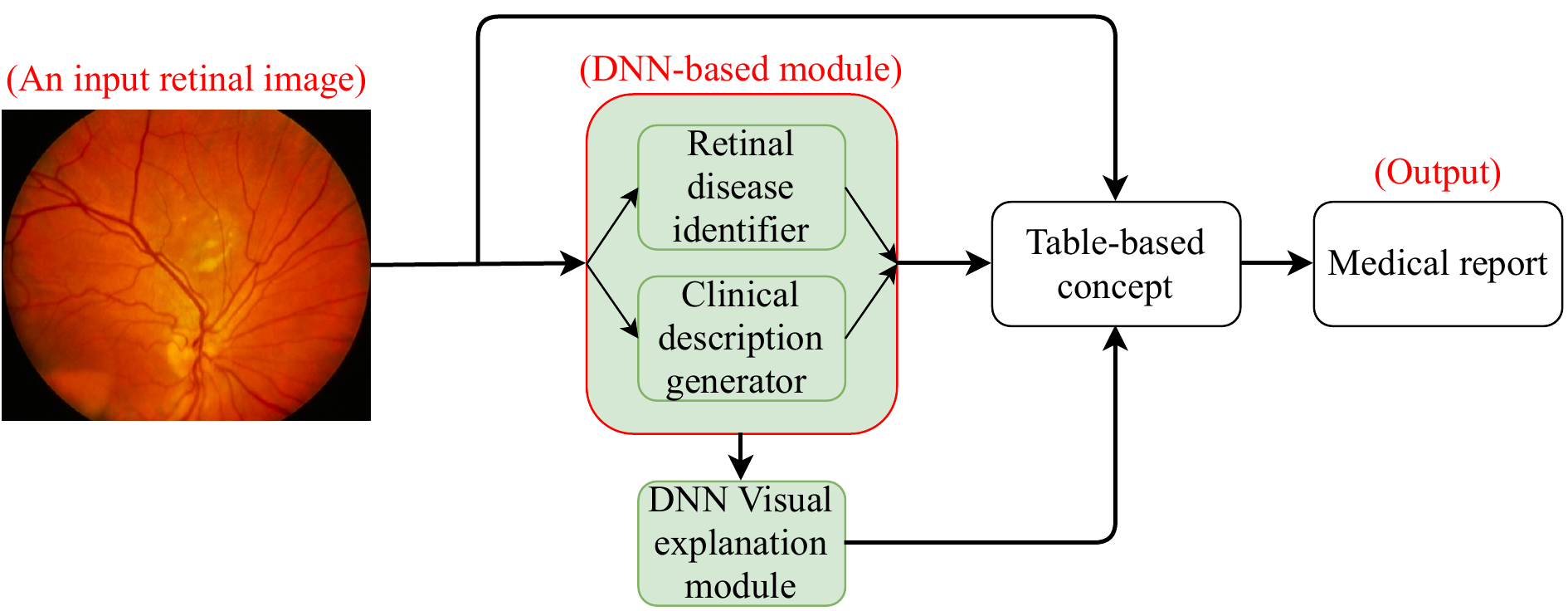}
\end{center}
\vspace{-0.6cm}
   \caption{This figure shows the proposed AI-based medical diagnosis method in the ophthalmology expert domain. It contains DNN-based and DNN Visual explanation modules. The DNN-based module is composed of two sub-modules, i.e., a retinal disease identifier and a clinical description generator reinforced by our proposed keyword-driven method, referring to our \textit{Methodology} section. The input of our method is a retinal image, and the output is a table-based \cite{zahalka2014towards} medical report. In Figure \ref{fig:figure1}, we shows how to exploit this AI-based method to improve the traditional retinal diseases treatment procedure. Note that, in this figure, DNN indicates deep neural networks.}
\vspace{-0.4cm}
\label{fig:figure100}
\end{figure}

\begin{figure}
\centering
  \includegraphics[width=1.0\linewidth]{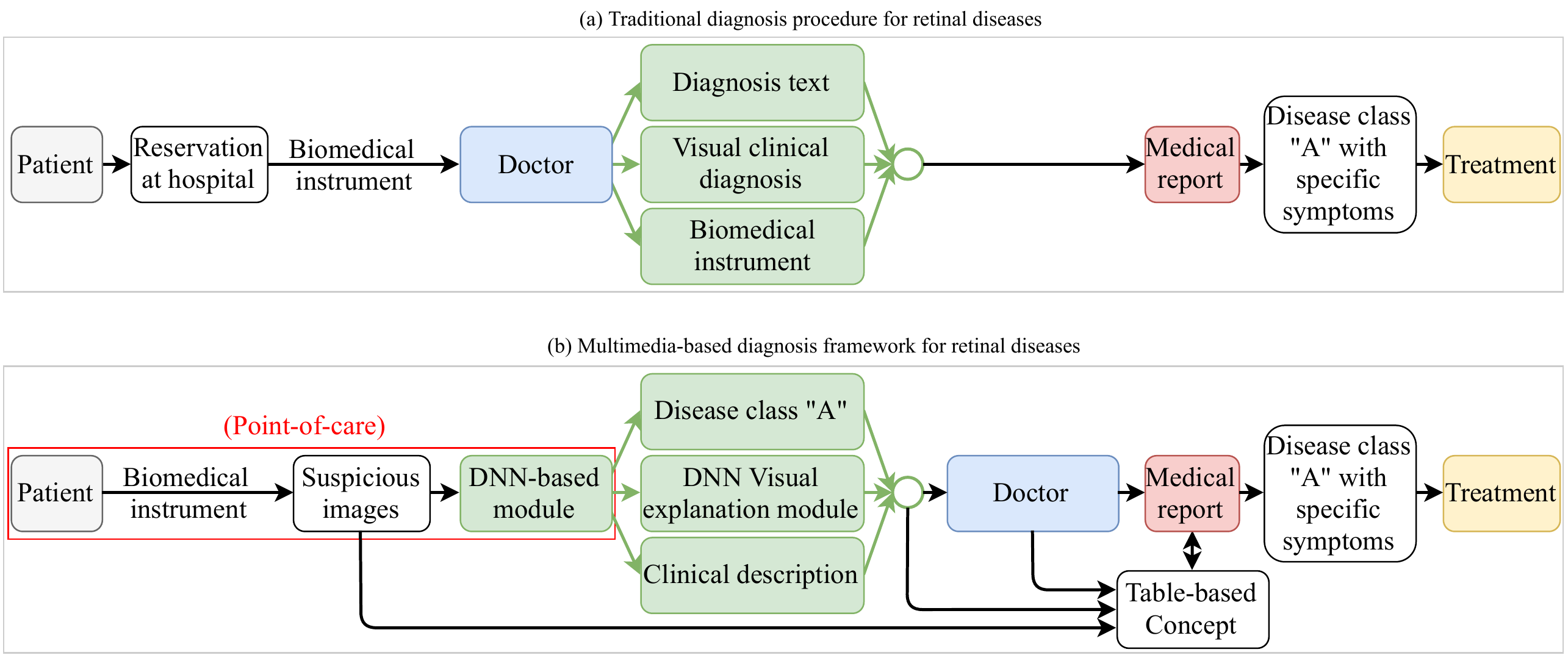}
    \vspace{-0.6cm}
  \caption{
    (a) is an existing traditional medical treatment procedure for retinal diseases \cite{tukey2014impact}. Typically, doctors have to handle most of the jobs in the traditional procedure. In (b), we incorporate the AI-based medical diagnosis method, referring to Figure \ref{fig:figure100}, in the traditional treatment procedure to improve the efficiency of (a), based on the point-of-care (POC) \cite{pai2012point} concept. In the proposed method, it mainly contains DNN-based and DNN visual explanation modules. The outputs of the DNN-based module are ``Disease class ``A'''' and ``Clinical description''. DNN visual explanation module visualize the information from the DNN-based module on the classification tasks. Please refer to our \textit{Methodology} section for a more detailed explanation. Note that DNN indicates deep neural networks in this figure.}
  \label{fig:figure1}
  \vspace{-0.4cm}
\end{figure}

In this paper, we propose an AI-based method for automatic medical report generation based on an input retinal image, as illustrated in Figure \ref{fig:figure100}. The proposed method intends to improve the traditional retinal disease diagnosis procedure, referring to Figure \ref{fig:figure1}, and help ophthalmologists increase diagnosis efficiency and accuracy. The main idea of this method is to exploit the deep learning based models, including an effective retinal disease identifier (RDI) and an effective clinical description generator (CDG), to automate part of the traditional treatment procedure. Then, the proposed method will make the diagnosis more efficient. 

To train our deep learning models and validate the effectiveness of our RDI and CDG, we introduce a new large-scale retinal disease image dataset, called DeepEyeNet (DEN). Besides, as ground truth, we provide a retinal image dataset manually labeled by ophthalmologists to qualitatively show that the proposed AI-based model is effective. The dataset helps us show the activation maps of our deep models are aligned with image features that are clinically recognized by ophthalmologists as linked with the identified disease. Our experimental results show that the proposed AI-based method is effective and successfully improves the traditional retinal disease treatment procedure. Our main contributions are summarized as follows:

\vspace{+1pt}
\noindent\textbf{Contributions.}
\vspace{-7pt}
\begin{itemize}
    \item To improve the traditional retinal disease treatment procedure and help ophthalmologists increase diagnosis efficiency and accuracy, we propose an AI-based method to generate medical reports for retinal images. In this method, we exploit the deep learning based models including an RDI and a CDG to automate part of the conventional treatment procedure.

    \item We propose a large-scale retinal disease image dataset, called DeepEyeNet (DEN) dataset, with 15,709 images to train our deep models and validate the effectiveness of the proposed RDI and CDG quantitatively.
    \item We provide another dataset with 300 retinal images labeled by ophthalmologists to qualitatively show our method is effective by visually confirming the activation maps of our models are aligned with image features clinically recognized by ophthalmologists.

\end{itemize}

\section{Related Work}
In this section, we divide the related works into retinal disease classification, image captioning, neural networks visual explanation, and retinal dataset comparison.

\noindent\textbf{2.1 Retinal Disease Classification}

Optical Coherence Tomography (OCT), Fluorescein Angiography (FA), and Color Fundus Photography (CFP) are the three most commonly used and important imaging methods for the diagnosis of retinal diseases~\cite{yanagihara2020methodological}. Optical Coherence Tomography (OCT) is a technology of emerging biomedical imaging, and it provides high-resolution and non-invasive real-time imaging of highly scattering tissues. That is, OCT images \cite{de2018clinically,lang2013retinal,gerckens2003optical} usually are used to show the structure of the retina. \cite{bagci2008thickness} have proposed an algorithm to segment and detect six different retinal layers, including Nerve Fiber Layer (NFL), Ganglion Cell Layer (GCL) + Inner Plexiform Layer (IPL), Inner Nuclear Layer (INL), Outer Plexiform Layer (OPL), Outer Nuclear Layer (ONL) + Photoreceptor Inner Segments (PIS), and Photoreceptor Outer Segments (POS), in OCT retinal images. Fluorescein Angiography (FA) has been used to realize the pathophysiologic course of Retinopathy of Prematurity (ROP) following intravitreal anti-Vascular Endothelial Growth Factor (Anti-VEGF) \cite{lepore2018follow}. Color Fundus photography (CFP) is a simple and cost-effective technology for trained medical professionals. Image preprocessing is one of the important issues in the automated analysis of CFP. The authors of \cite{youssif2006comparative} have proposed a method to reduce the vignetting effect caused by non-uniform illumination of a retinal image. In this work, we mainly exploit the DNN-based methods \cite{he2016deep,simonyan2014very,yang2020net} to further investigate the retinal disease classification~\cite{yang2018novel} toward a multi-label task combining language information.

\noindent\textbf{2.2 Image Captioning}

Recently, computer vision researchers have proposed a new task, image captioning, and \cite{karpathy2015deep,vinyals2015show,fang2015captions} are early works. In \cite{karpathy2015deep}, the proposed model can embed visual and language information into a common multimodal space. The authors of \cite{fang2015captions} exploit a natural language model to combine a set of possible words, which are related to several small parts of the image, and then generate the caption of the given image. The authors of \cite{vinyals2015show} use CNN to extract the image feature and use it as the input at the first time step of the RNN to generate the caption of the input image. The authors of \cite{gao2019deliberate} propose a new deliberate residual attention network for image captioning. The layer of first-pass residual-based attention prepares the visual attention and hidden states for generating a preliminary version of the captions, while the layer of second-pass deliberate residual-based attention refines them. Since the second-pass is based on the global features captured by the hidden layer and visual attention in the first-pass, their method has the potentials to generate better captions. In \cite{liu2017improved}, the authors mention that existing image captioning models are usually trained via maximum likelihood estimation. However, the log-likelihood score of some captions cannot correlate well with human assessments of quality. Standard syntactic text evaluation metrics, such as METEOR \cite{banerjee2005meteor}, BLEU \cite{papineni2002bleu}, and ROUGE \cite{lin2004rouge}, are also not well correlated. The authors of \cite{liu2017improved} show how to use a policy gradient method to optimize a linear combination of CIDEr \cite{vedantam2015cider} and SPICE \cite{anderson2016spice}. In \cite{hendricks2016generating}, the authors propose a method that focuses on discriminating properties of the visible object, jointly predicts a class label, and explains why the predicted label is proper for a given image. Through a loss function based on reinforcement learning and sampling, their model learns to generate captions. According to \cite{vinyals2015show,karpathy2015deep,gao2019deliberate}, existing image captioning models are only able to generate the rough description for a given image. So, in this work, 
we exploit keywords to make our CDG have better reasoning ability.

\begin{table*}
    \caption{Summary of available retinal datasets. Based on this table, we find our proposed DEN is much larger than the other retinal image datasets. It contains three types of labels including the name of the disease, keywords, and clinical description. Most of the retinal dataset only contains image data, and the dataset size is not large. Note that `` Text*'' denotes clinical description and keywords, referring to our \textit{Dataset Introduction and Analysis} section. `` Text'' denotes only clinical description. So, our DEN is unique.}
    \vspace{-0.6cm}
\begin{center}
\scalebox{1.0}{
    \begin{tabular}{|P{3.0cm}|P{2.75cm}|P{3.5cm}|P{2.75cm}|P{3.0cm}|}
    \hline
    \textbf{Name of Dataset} & \textbf{Field of View} & \textbf{Resolution} & \textbf{Data Type} & \textbf{Number of Images}\\ \hline
    VICAVR \cite{vazquez2013improving} & $45^{\circ}$ & $768*584$ & Image & 58\\ \hline
    VARIA \cite{ortega2009retinal} & $20^{\circ}$ & $768*584$ & Image & 233\\ \hline
    STARE \cite{hoover2003locating} & $\approx 30^{\circ}-45^{\circ}$ & $700*605$ & Image + Text & 397\\ \hline
    CHASE-DB1 \cite{fraz2012ensemble} & $\approx 25^{\circ}$ & $999*960$ & Image & 14\\ \hline
    RODREP \cite{adal2015accuracy} & $45^{\circ}$ & $2000*1312$ & Image & 1,120\\ \hline
    HRF \cite{odstrvcilik2009improvement} & $45^{\circ}$ & 3504*2336 & Image & 45\\ \hline
    e-ophtha \cite{decenciere2013teleophta} & $\approx 45^{\circ}$ & $2544*1696$ & Image & 463\\ \hline    
    ROC \cite{niemeijer2010retinopathy} & $\approx 30^{\circ}-45^{\circ}$ & $768*576 - 1386*1391$ & Image & 100\\ \hline
    REVIEW \cite{al2008reference} & $\approx 45^{\circ}$ & $1360*1024 - 3584*2438$ & Image & 14\\ \hline
    ONHSD \cite{computing2012understanding} & $45^{\circ}$ & $640*480$ & Image & 99\\ \hline   
    INSPIRE-AVR \cite{niemeijer2011inspire} & $30^{\circ}$ & $2392*2048$ & Image & 40\\ \hline    
    DIARETDB1 \cite{kauppi2007diaretdb1} & $50^{\circ}$ & $1500*1152$ & Image + Text & 89\\ \hline
    DIARETDB0 \cite{kauppidiaretdb0} & $50^{\circ}$ & $1500*1152$ & Image & 130\\ \hline  
    MESSIDOR \cite{decenciere2014feedback} & $45^{\circ}$ & $1440*960 - 2304*1536$ & Image + Text & 1,200\\ \hline
    Drishti-GS \cite{sivaswamy2014drishti} & $\approx 25^{\circ}$ & $2045*1752$ & Image & 101\\  \hline
    FIRE \cite{hernandez2017fire} & $45^{\circ}$ & $2912*2912$ & Image & 129\\  \hline  
    DRIONS-DB \cite{carmona2008identification} & $\approx 30^{\circ}$ & $600*400$ & Image & 110\\  \hline 
    IDRiD \cite{porwal2018indian} & $50^{\circ}$ & $4288*2848$ & Image & 516\\  \hline 
    DRIVE \cite{staal2004ridge} & $45^{\circ}$ & $565*584$ & Image & 40\\ \hline
    \textbf{DeepEyeNet (DEN)} & $\approx \textbf{30}^{\circ}-\textbf{60}^{\circ}$ & \textbf{various} & \textbf{Image + Text*} & \textbf{15,709}\\ \hline 
    
    \end{tabular}}

    \label{table:table103}
\vspace{-0.8cm}
\end{center}
\end{table*}

\noindent\textbf{2.3 Neural Networks Visual Explanation}

There are some popular CNN visualization tools, \cite{zhou2015cnnlocalization,selvaraju2017grad,yang2019causal}. The authors of \cite{zhou2015cnnlocalization} have proposed a technique, called Class Activation Mapping (CAM), for CNN. It makes classification-trained CNN learn how to perform the task of object localization, without using a bounding box. In our previous works~\cite{liu2018synthesizing, yang2018auto}, we exploit class activation maps to visualize the predicted class scores on retinal images, highlighting the discriminative object parts which are detected by the CNN. In \cite{selvaraju2017grad}, the authors have proposed the other similar features visualization tool, called Gradient-weighted Class Activation Mapping (Grad-CAM), for making a CNN-based model transparent by producing visual explanations of features. The authors of \cite{zeiler2014visualizing} introduce a CNN visualization technique that gives insight into the operation of the classifier and the function of intermediate feature layers. These visualizations allow us to find architectures of CNN models. The authors of \cite{chattopadhay2018grad} propose a generalized method, Grad-CAM++, based on Grad-CAM. The Grad-CAM++ method provides better visual explanations of CNN model predictions than Grad-CAM, in terms of better object localization~\cite{yang2020wavelet} and occurrences explanation of multiple object instances in a single image. In \cite{li2018tell}, the authors propose another method different from the above methods which are trying to explain the network. They build up an end-to-end model to provide supervision directly on the visual explanations. Furthermore, the authors validate that the supervision can guide the network to focus on some expected regions. The aforementioned is more related to image data only visualization. The authors of \cite{zahalka2014towards,rooij2012efficient,huang2020query} have proposed some methods for the multimedia data, such as text and images, visualization. In \cite{zahalka2014towards}, the authors introduce five popular multimedia visualization concepts, including basic grid, similarity space, similarity-based, spreadsheet, and thread-based concepts. In this work, we exploit CAM to visually show that the activation maps of our deep models are aligned with image features that are clinically recognized by ophthalmologists as linked with the identified disease. In addition, we use a table-based concept, similar to the static spreadsheet concept, to visualize our medical report.

\noindent\textbf{2.4 Retinal Dataset Comparison}

Retinal disease research already has long history and many retinal datasets have been proposed, such as \cite{staal2004ridge,porwal2018indian,carmona2008identification,hernandez2017fire,sivaswamy2014drishti,decenciere2014feedback,kauppidiaretdb0,kauppi2007diaretdb1,niemeijer2011inspire,computing2012understanding,al2008reference,niemeijer2010retinopathy,decenciere2013teleophta,odstrvcilik2009improvement,adal2015accuracy,fraz2012ensemble,hoover2003locating,ortega2009retinal,vazquez2013improving}. The DRIVE dataset \cite{staal2004ridge} contains 40 retina images which are obtained from a diabetic retinopathy screening program in the Netherlands. These 40 images have been divided into a half training set and a half test set. For the training images, a single manual segmentation of the vasculature is available. For the test cases, two manual segmentations are available.
The IDRiD dataset \cite{porwal2018indian} is a dataset for retinal fundus image consisting of 516 images. The authors of IDRiD dataset provide ground truths associated with the signs of Diabetic Macular Edema (DME) and Diabetic Retinopathy (DR) and normal retinal structures given below and described as follows: (i) Pixel level labels of typical DR lesions and optic disc; (ii) Image level disease severity grading of DR, and DME; (iii) Optic disc and fovea center coordinates. The DRIONS-DB dataset \cite{carmona2008identification} consists of 110 color digital retinal images, and it contains several visual characteristics, such as cataract (severe or moderate), light artifacts, some of the rim blurred or missing, moderate peripapillary atrophy, concentric peripapillary atrophy/artifacts, and strong pallor distractor. The FIRE dataset \cite{hernandez2017fire} consists of 129 retinal images forming 134 image pairs, and image pairs are split into three different categories depending on their characteristics. The Drishti-GS dataset \cite{sivaswamy2014drishti} contains 101 images, and it is divided into 50 training and 51 testing images. The MESSIDOR dataset \cite{decenciere2014feedback} has 1200 eye fundus color numerical images. Although the dataset contains a medical diagnosis for each image, there is no manual annotation, such as lesions contours or position, on the images. The DIARETDB0 dataset \cite{kauppidiaretdb0} consists of 130 color fundus images of which 20 are normal, and 110 contain signs of the DR. The DIARETDB1 dataset \cite{kauppi2007diaretdb1} consists of 89 color fundus images of which 84 contain at least mild non-proliferative signs of the DR, and five are considered as normal which do not contain any signs of the DR. The INSPIRE-AVR dataset \cite{niemeijer2011inspire} has 40 colorful images of the vessels and optic disc and an arterio-venous ratio reference standard. The ONHSD  dataset \cite{computing2012understanding} has 99 retinal images and it is mainly used for the segmentation task. The REVIEW dataset \cite{al2008reference} consists of 14 images, and it is also mainly used for the segmentation task. The ROC dataset \cite{niemeijer2010retinopathy} aims to help patients with diabetes through improving computer-aided detection and diagnosis of DR. The e-ophtha \cite{decenciere2013teleophta} is a dataset of color fundus images specially designed for scientific research in DR. The HRF dataset \cite{odstrvcilik2009improvement} contains at the moment 15 images of healthy patients, 15 images of patients with DR and 15 images of glaucomatous patients. Also, binary gold standard vessel segmentation images are available for each image. The RODREP dataset \cite{adal2015accuracy} contains repeated 4-field color fundus photos (1120 in total) of 70 patients in the DR screening program of the Rotterdam Eye Hospital. The CHASE-DB1 dataset \cite{fraz2012ensemble} is mainly used for retinal vessel analysis, and it contains 14 images. The STARE dataset \cite{hoover2003locating} has 397 images and it is used to develop an automatic system for diagnosing diseases of the human eye. The VARIA \cite{ortega2009retinal} is a dataset of retinal images used for authentication purposes, and it includes 233 images from 139 different individuals. The VICAVR dataset \cite{vazquez2013improving} includes 58 images, and it is used for the computation of the ratio of A/V, (Artery/Vein). In this work, we propose a large-scale retinal images dataset, DeepEyeNet (DEN), to train our deep learning based models and validate our RDI and CDG. For convenience, we summarize the above retinal datasets in Table \ref{table:table103}.

\section{Dataset Introduction and Analysis}
\begin{figure}[t!]
\begin{center}
\includegraphics[width=1.0\linewidth]{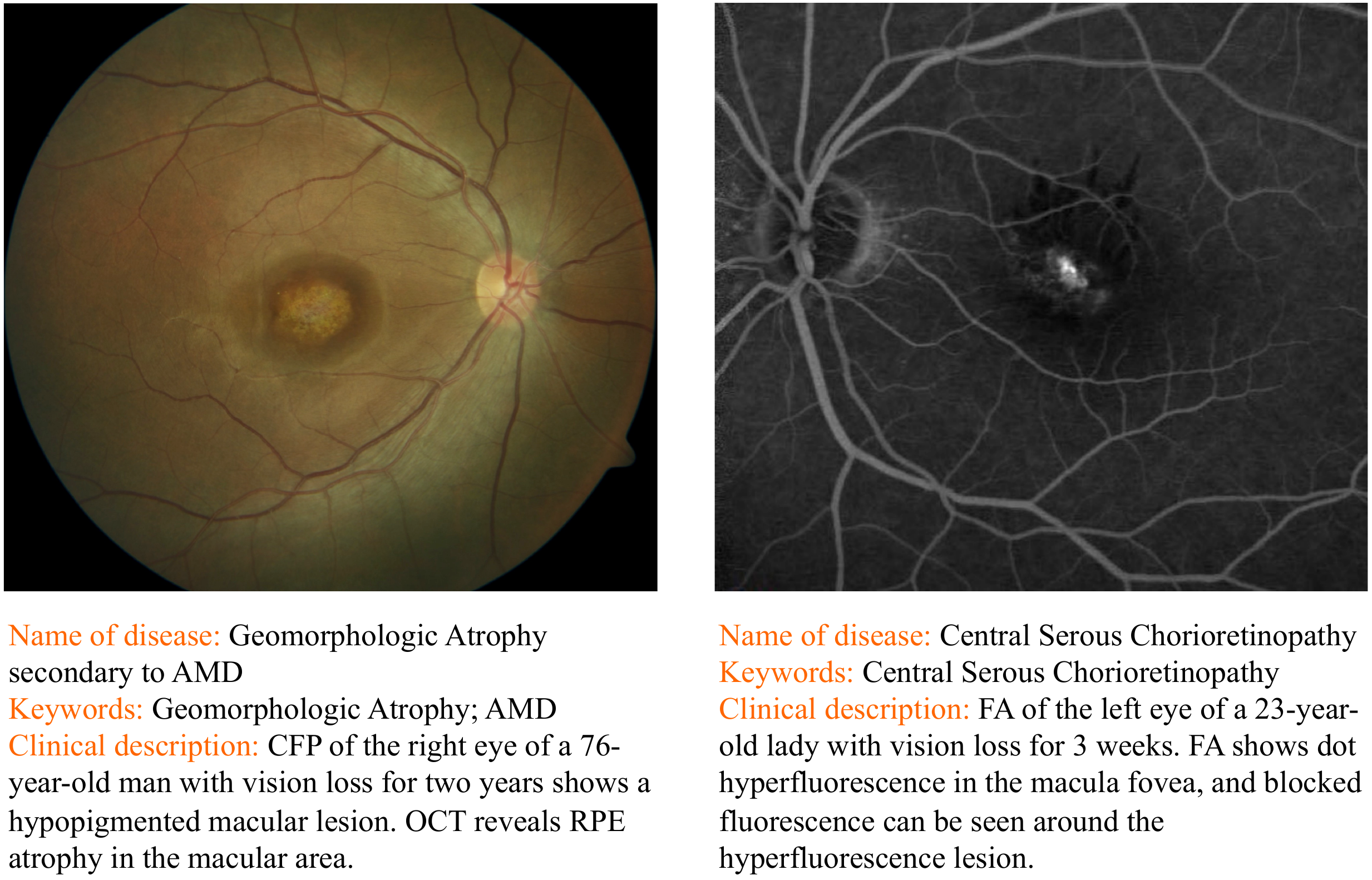}
\end{center}
\vspace{-0.6cm}
   \caption{Examples from our DEN dataset. Each image has three labels including the name of the disease, keywords, and clinical description. Note that ophthalmologists define all the labels.}
\vspace{-0.2cm}
\label{fig:figure2}
\end{figure}

\begin{figure}[t!]
\begin{center}
\includegraphics[width=1.0\linewidth]{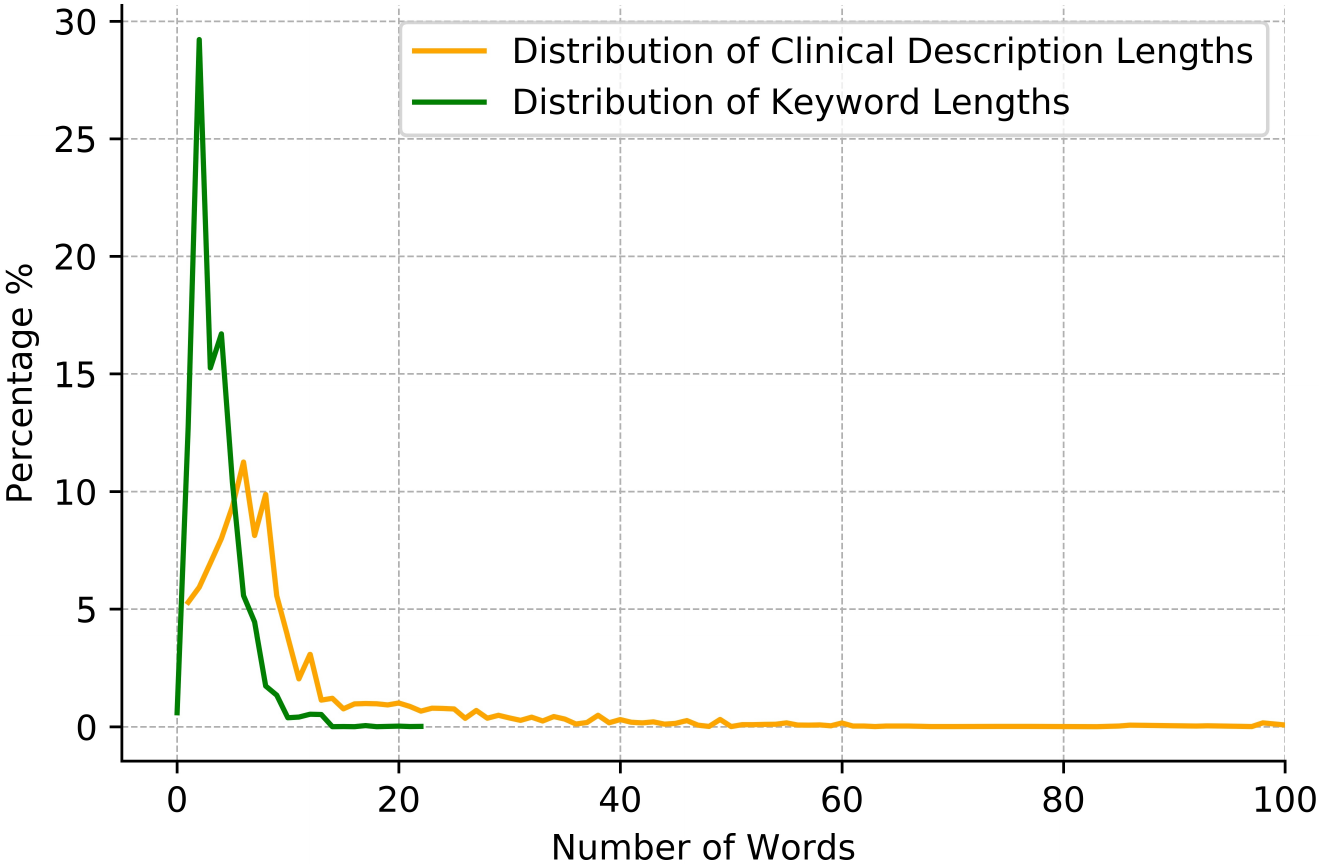}
\end{center}
\vspace{-0.6cm}
   \caption{This figure shows the word length distribution of the keyword and clinical description labels. Based on the figure, the word length in our DEN dataset is mainly between $5$ and $10$ words.}
\vspace{-0.2cm}
\label{fig:figure3}
\end{figure}

\begin{figure}[t!]
\begin{center}
\includegraphics[width=1.0\linewidth]{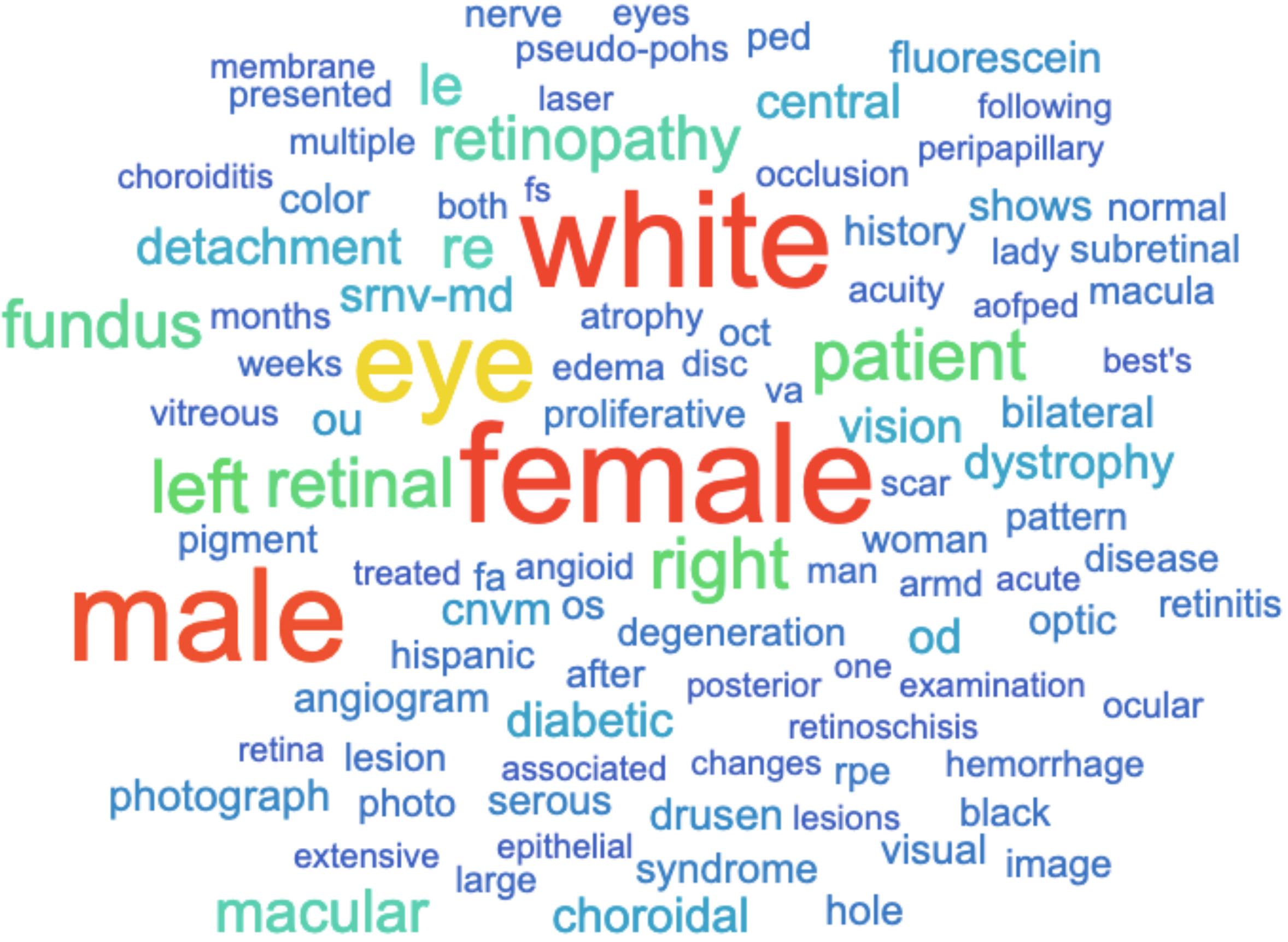}
\end{center}
\vspace{-0.6cm}
  \caption{The figure represents Venn-style word cloud for clinical description labels. Note that the word size indicates the normalized counts. Based on this figure, we can see there are specific abstract concepts, which makes image captioning algorithms more difficult to generate descriptions with good quality.}
\vspace{-0.2cm}
\label{fig:figure4}
\end{figure}

In this section, we start to describe our proposed DEN dataset in terms of types of retinal images and labels and some statistics of the dataset. Note that some of our group members are experienced ophthalmologists and they help us build the proposed DEN dataset sorted by 265 unique retinal symptoms from the clinical definition and their professional domain knowledge. In our proposed DEN dataset, there are two types of retinal images, grey scale FA and colorful CFP. 
The total amount of images is 15,709, including 1,811 FA and 13,898 CFP. As with most of the large-scale datasets for deep learning research, we create standard splits, separating the whole dataset into $60\%/20\%/20\%$, i.e., $9425/3142/3142$, for training/validation/testing, respectively. 
Each retinal image has three corresponding labels including the name of the disease, keywords, and clinical description. For the total number of retinal diseases, the dataset contains 265 different retinal diseases including the common and non-common. For the keyword and clinical description, it contains 15,709 captions and 15,709 keywords labels. Keyword label denotes important information in the diagnosis process. Clinical description label represents the corresponding caption of a given retinal image. Note that all the labels are defined by retina specialists or ophthalmologists. To better understand our dataset, we show some data examples from the DEN dataset in Figure \ref{fig:figure2}. Also, in Figure \ref{fig:figure3}, we show the word length distribution of the keyword and clinical description labels. Based on Figure \ref{fig:figure3}, we observe the longest word length in our dataset is more than $15$ words for keywords and $50$ words for clinical descriptions. Note that the longest word length of existing datasets for natural image captioning or VQA \cite{antol2015vqa,chen2015microsoft,lin2014microsoft,huang2019assessing,huang2017robustnessMS} is only around $10$ words. It implies that our proposed dataset is challenging. Additionally, we provide the Venn-style word cloud visualization results clinical description labels, referring to Figure \ref{fig:figure4}. Based on Figure \ref{fig:figure4}, in clinical description labels, we see there are specific abstract concepts, which makes the dataset more challenging.

\begin{figure}[ht]
\begin{center}
\includegraphics[width=1.0\linewidth]{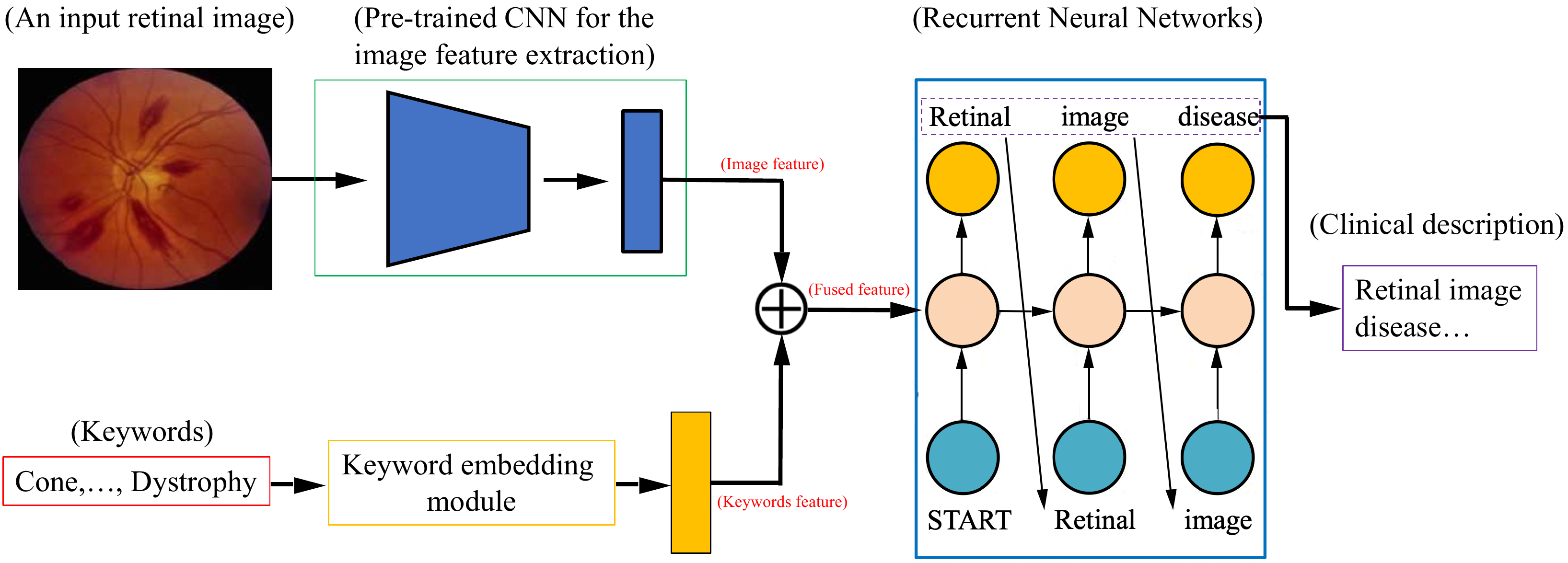}
\end{center}
\vspace{-0.6cm}
   \caption{This figure conceptually depicts the clinical description generator with our proposed keyword-driven method. In our clinical description generator, we exploit a pre-trained CNN model to extract the retinal image feature. So, the CNN model is a so-called image encoder. Then, we use an LSTM model, i.e., recurrent neural networks (RNN), as a decoder to generate a word at each time step. Finally, all of the collected words will form a clinical description.}
\vspace{-0.2cm}
\label{fig:figure6}
\end{figure}

\section{Methodology}
In this section, we start to describe the proposed AI-based method for automatic medical report generation. The proposed method is mainly composed of the DNN-based module and DNN visual explanation module.

\noindent\textbf{4.1 DNN-based Module}

The DNN-based module contains two components, i.e., a retinal disease identifier (RDI) and a clinical description generator (CDG). We introduce them in the following subsections. Note that we hypothesize an effective RDI and effective CDG help improve the conventional retinal disease treatment procedure and help ophthalmologists increase diagnosis efficiency and accuracy.

\noindent\textbf{Retinal Disease Identifier (RDI).}
To identify retinal diseases, in our RDI sub-module, we provide two types of deep learning models based on \cite{he2016deep,simonyan2014very}, pre-trained on ImageNet, and then trained on the proposed DEN dataset. From the lower level feature perspective, such as color, most of the medical images, e.g., radiology images of the chest, are mainly grey-scale \cite{laserson2018textray} but retinal images are mainly colorful in our dataset. Using the ImageNet pre-trained at least helps extract the better lower level features information. So, in this case, we expect that pre-training on ImageNet can improve model performance.

\noindent\textbf{Clinical Description Generator (CDG).}
To generate the clinical description for an input retinal image, we use a pre-trained CNN-based model, such as MobileNetV2, VGG16, VGG19, or InceptionV3, as our image feature encoder and a Long Short-term Memory (LSTM) as our decoder to generate text, referring to Figure \ref{fig:figure6}. When we try to generate the clinical description by the LSTM unit, we incorporate the beam-search mechanism to get the better final output description. In ophthalmological practice, commonly existing keywords, with unordered nature, help ophthalmologists create medical reports. Inspired by this, we exploit keywords to reinforce our CDG sub-module. As shown in Figure \ref{fig:figure6}, we use a keyword embedding module, such as bag of words, to encode our keyword information. Note that when keywords are used to reinforce CDG, it means we will have two types of input features, i.e., image and text features. In this work, we use the average method to fuse these two types of features, referring to Figure \ref{fig:figure6}.

\noindent\textbf{4.2 DNN Visual Explanation Modules}

There are some existing DNN visual explanation methods, such as \cite{zhou2015cnnlocalization,selvaraju2017grad,hu2019silco, liu2020interpretable}. The authors of \cite{zhou2015cnnlocalization} have proposed a technique, called Class Activation Mapping (CAM), for CNN. It makes classification-trained CNN learn how to perform the task of object localization, without using a bounding box. Furthermore, they exploit class activation maps to visualize the predicted class scores on a given image, highlighting the discriminative object parts which are detected by the CNN. To improve the conventional retinal disease treatment procedure, we incorporate the DNN visual explanation module in our proposed AI-based method. Also, we exploit this module to help verify the effectiveness of the method, referring to our \textit{Experiments} section.

\begin{table}[t!]
\caption{
This table shows the quantitative results of different RDI models based on our DEN. The RDI model based on \cite{simonyan2014very} with ImageNet pre-training has the best performance. ``Pre-trained'' indicates the model is initialized from the pre-trained weights of ImageNet. ``Random init'' means the model's weights are initialized randomly. Prec@k indicates how often the ground truth label is within the top $k$ ranked labels after the \textit{softmax} layer. We investigate Prec@1 and Prec@5 due to the need to shortlist candidates of diseases in real-world scenarios. Note that since we have $265$ retinal disease candidates and limited training data, it is hard to have good performance in the sense of Prec@1. The situation of limited data is common in medicine.}
\vspace{-0.3cm}
\centering
\scalebox{0.83}{
\begin{tabular}{|l|llll|}
\hline
\multirow{3}{*}{Model} & \multicolumn{4}{l|}{Precision}                                                          \\ \cline{2-5} 
                      & \multicolumn{2}{l}{Pre-trained}                       & \multicolumn{2}{l|}{Random init} \\ \cline{2-5} 
                      & Prec@1         & \multicolumn{1}{l|}{Prec@5}         & Prec@1         & Prec@5         \\ \hline
He, et al. \cite{he2016deep}              & 37.09 & \multicolumn{1}{l|}{63.36} & \textbf{36.60}          & 62.87          \\
Simonyan, et al. \cite{simonyan2014very}                 & \textbf{54.23}          & \multicolumn{1}{l|}{\textbf{80.75}}          & 35.93          & \textbf{73.73}          \\
Jing, et al. \cite{jing2018automatic}                 & 32.72          & \multicolumn{1}{l|}{63.75}          & 29.11          & 60.68          \\
\hline
\end{tabular}}

\label{table:table1}
\vspace{-0.2cm}
\end{table}

\noindent\textbf{4.3 Medical Report Generation}

\begin{figure*}[t!]
\begin{center}
\includegraphics[width=1.0\linewidth]{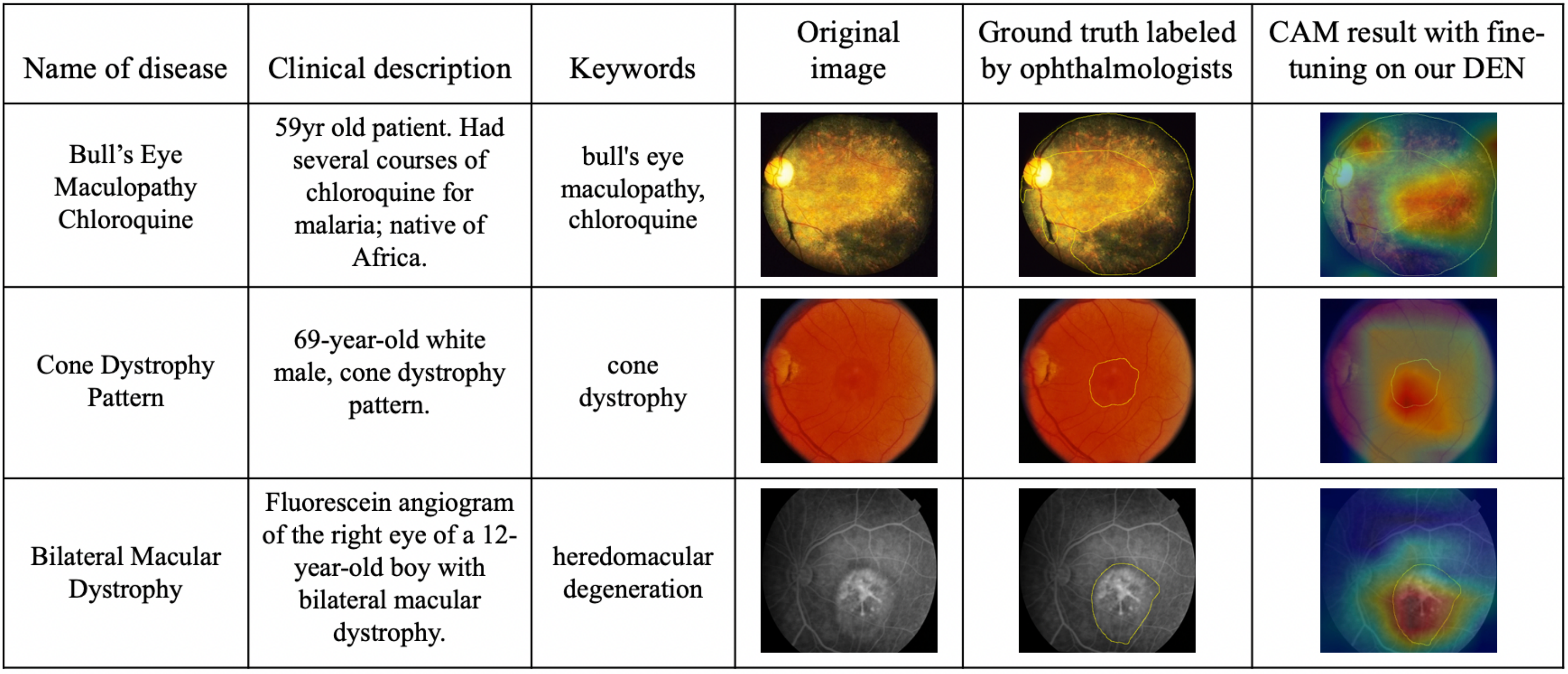}
\end{center}
\vspace{-0.6cm}
   \caption{This figure shows the medical reports based on the table-based concept \cite{zahalka2014towards}. Since retinal diseases may have some implicit common property or relation, we can put the diseases with the common property or relation together on the table. The table-based medical report intends to help ophthalmologists get more insights.}
\label{fig:figure12}
\end{figure*}

According to \cite{zahalka2014towards,rooij2012efficient}, proper multimedia data visualization helps people get insight from the data efficiently. In some sense, we can say that multimedia visualization is a way to visually arrange multimedia data, and it sometimes even helps people get a deeper understanding and extra information from the visualized data. In this work, it contains five multimedia data, including the name of the disease, keyword, clinical description, retinal image, and CAM result image. So, we exploit the table-based concept, which is similar to the static spreadsheet-based concept \cite{zahalka2014towards}, to visualize our medical report, referring to Figure \ref{fig:figure12}. The medical report visualization intends to help ophthalmologists get the insight from the above image and text data efficiently and also increase the diagnostic accuracy.

\begin{figure*}[t!]
\begin{center}
\includegraphics[width=1.0\linewidth]{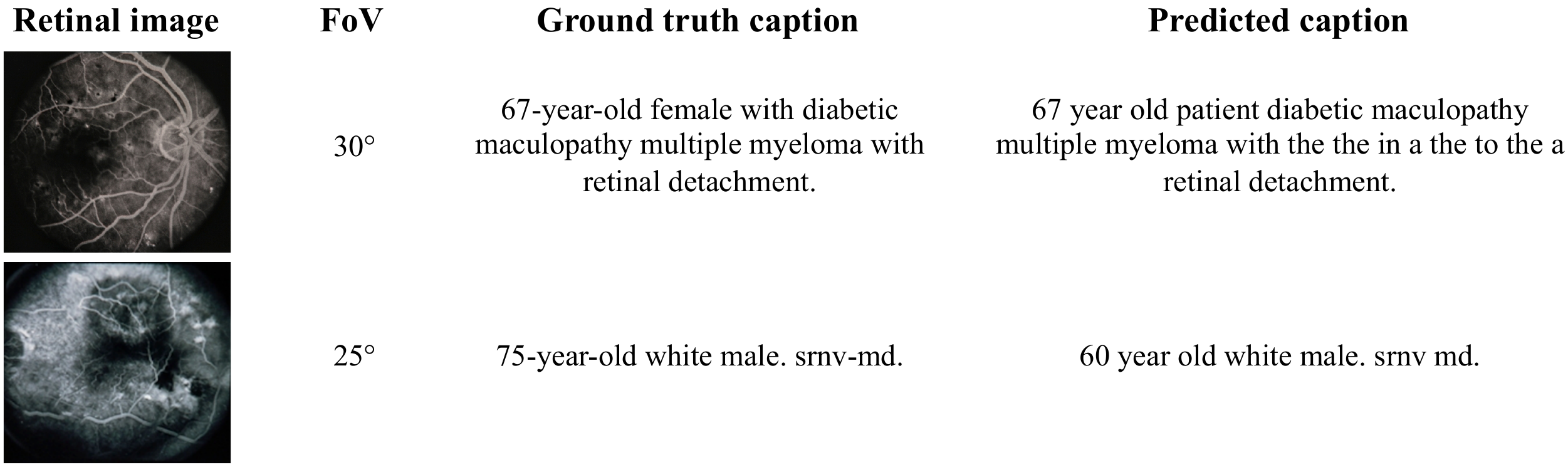}
\end{center}
\vspace{-0.60cm}
  \caption{This figure shows some generated results by our clinical description generator. Based on this figure, we know that our models can generate meaningful clinical descriptions for ophthalmologists. Note that, in practice, ``age'' and ``gender'' are hard to be generated correctly by automatic algorithms. The first row with correct ``age'' prediction is just a special case.}
\vspace{-0.30cm}
\label{fig:figure11}
\end{figure*}

\section{Experiments}
In this section, we compare our proposed method to baselines for verifying the effectiveness based on the assumption described in our \textit{Methodology} section.

\noindent\textbf{5.1 Retinal Disease Identifier (RDI) Verification}

In our experiment, we try to show that the RDI model with ImageNet pre-training is better than the RDI model without ImageNet pre-training, i.e., our baseline. We exploit the ImageNet-pre-trained DNN-based deep model and non-ImageNet-pre-trained DNN-based deep model with different architectures to do fine-tuning on DEN. For empirical reasons, we use two recipes to train different models. For the RDI model based on \cite{he2016deep}, we start with a learning rate of $0.1$ and decay it $5$ times for every $50$ epoch. For the RDI model based on \cite{simonyan2014very}, we start with a learning rate of $0.001$ and decay it $5$ times for every $50$ epoch. According to the evaluation results in Table \ref{table:table1}, we find that the RDI model based on \cite{simonyan2014very} with ImageNet pre-training has better performance than others. We conjecture that RDI models based on \cite{he2016deep,jing2018automatic} may be too complicated for the proposed DEN dataset. Although DEN is a large-scale dataset from the retinal field perspective, the number of training images is still not enough for very deep models. Note that our proposed DEN dataset has $265$ classes, including common and non-common retinal diseases or symptoms, and only $8512$ training images, so it is not easy to achieve high Prec@1 accuracy for human doctors and AI machines. That is one of the reasons why we investigate both Prec@1 and Prec@5. Also, reporting Prec@5 accuracy is more appropriate from the real-world scenario perspective.

\begin{table*}[t!]
\caption{This table shows the evaluation results of our keyword-driven and non-keyword-driven clinical description generators (CDGs). Note that we highlight the best scores of keyword-driven and non-keyword-driven generators in each column, respectively. ``w/o'' denotes non-keyword-driven baseline generators, and ``w/'' denotes our proposed keyword-driven generators. ``BLEU-avg'' denotes the average score of BLEU-1, BLEU2, BLEU-3, and BLEU-4. Note that the model based on ``Jing, et al. \cite{jing2018automatic}'' has the best performance among all the non-keyword-driven models, and the keyword-driven model based on ``Jing, et al. \cite{jing2018automatic}'' also has the best performance among all the models. 
All the keyword-driven models, with the average feature fusion method, are superior to the non-keyword-driven models. So, using keywords to reinforce the CDGs is effective.}
\vspace{-0.2cm}
\centering
\scalebox{1.0}{
\begin{tabular}{|c|l|l|l|l|l|l|l|l|}
\hline
\multicolumn{2}{|c|}{Model}                  & BLEU-1 & BLEU-2 & BLEU-3 & BLEU-4 & BLEU-avg & CIDEr & ROUGE \\ \hline
\multirow{2}{*}{Karpathy, et al. \cite{karpathy2015deep}}           & w/o & 0.081  & 0.031  & 0.009  & 0.004  & 0.031    & 0.117 & 0.134 \\ \cline{2-9}
                                    & w/ &  \textbf{0.169}     & \textbf{0.103}   & \textbf{0.060}      & \textbf{0.017}      & \textbf{0.087}      & \textbf{0.120}            & \textbf{0.202} \\ \hline
\multirow{2}{*}{Vinyals, et al. \cite{vinyals2015show}}              & w/o & 0.054  & 0.018  & 0.002  & 0.001  & 0.019    & 0.056 & 0.083 \\ \cline{2-9} 
                                    & w/ &  \textbf{0.144}     & \textbf{0.092}   & \textbf{0.052}      & \textbf{0.021}      & \textbf{0.077}      & \textbf{0.296}            & \textbf{0.197} \\ \hline
\multirow{2}{*}{Jing, et al. \cite{jing2018automatic}}              & w/o & 0.130  & 0.083  & 0.044  & 0.012  & 0.067    & 0.167 & 0.149 \\ \cline{2-9} 
                                    & w/ &  \textbf{0.184}     & \textbf{0.114}      & \textbf{0.068}       & \textbf{0.032}    & \textbf{0.100}     & \textbf{0.361}            & \textbf{0.232} \\ \hline
\multirow{2}{*}{Li, et al. \cite{li2019knowledge}}        & w/o & 0.111  & 0.060  & 0.026  & 0.006  & 0.051    & 0.066 & 0.129 \\ \cline{2-9} 
                                    & w/ & \textbf{0.181}    & \textbf{0.107}  & \textbf{0.062}   & \textbf{0.032}   & \textbf{0.096}   & \textbf{0.453}   & \textbf{0.230} \\ \hline
\end{tabular}}
\label{table:table200}
\end{table*}

\noindent\textbf{5.2 Clinical Description Generator (CDG) Verification}

In \cite{huang2017vqabq,huang2019novel,huang2017robustness}, the authors mention that the evaluation of image description generators is very subjective and there is no such thing as the most proper metric to evaluate the text-to-text similarity. Different text-to-text similarity metrics have different properties, so we exploit six commonly used metrics, including BLEU-1, BLEU-2, BLEU-3, BLEU-4 \cite{papineni2002bleu}, ROUGE \cite{lin2004rouge}, and CIDEr \cite{vedantam2015cider}, to evaluate generated results by our CDG. Table \ref{table:table200} contains the evaluation results of our CDGs based on the above six different text-to-text similarity metrics. All CDG modules with the keyword-driven method have better performance than the non-keyword-driven CDGs, i.e., our baseline. It implies that using keywords to reinforce the CDGs is effective. Based on Table \ref{table:table200} and \cite{jing2018automatic,wang2018tienet}, we find that the evaluation score of the medical image captioning, based on the above commonly used text evaluation metrics, is much lower than the evaluation score of the natural image captioning. One reason is that, typically, the length of the medical image caption is much longer than the natural image caption. Also, the medical image caption has more abstract words or concepts than the natural image caption. These abstract words/concepts will make algorithms difficult to generate correct captions. The other possible reason is that the innate property of the commonly used text-to-text similarity metrics \cite{huang2017vqabq,huang2019novel,huang2017robustness} makes this happen. In addition, in Figure \ref{fig:figure11}, we show some generated clinical description results. Based on Figure \ref{fig:figure11}, we find that although our CDG module cannot always generate correct ``age'' or ``gender'', the models are capable of generating correct descriptions to important characteristics for retinal images.

Based on the assumption mentioned in our \textit{Methodology} section, subsection 5.1, and subsection 5.2, we have shown the proposed AI-based method is quantitatively effective.

\noindent\textbf{5.3 Evaluation by DNN Visual Explanation Module}

\begin{figure}[t!]
\begin{center}
   \includegraphics[width=1.0\linewidth]{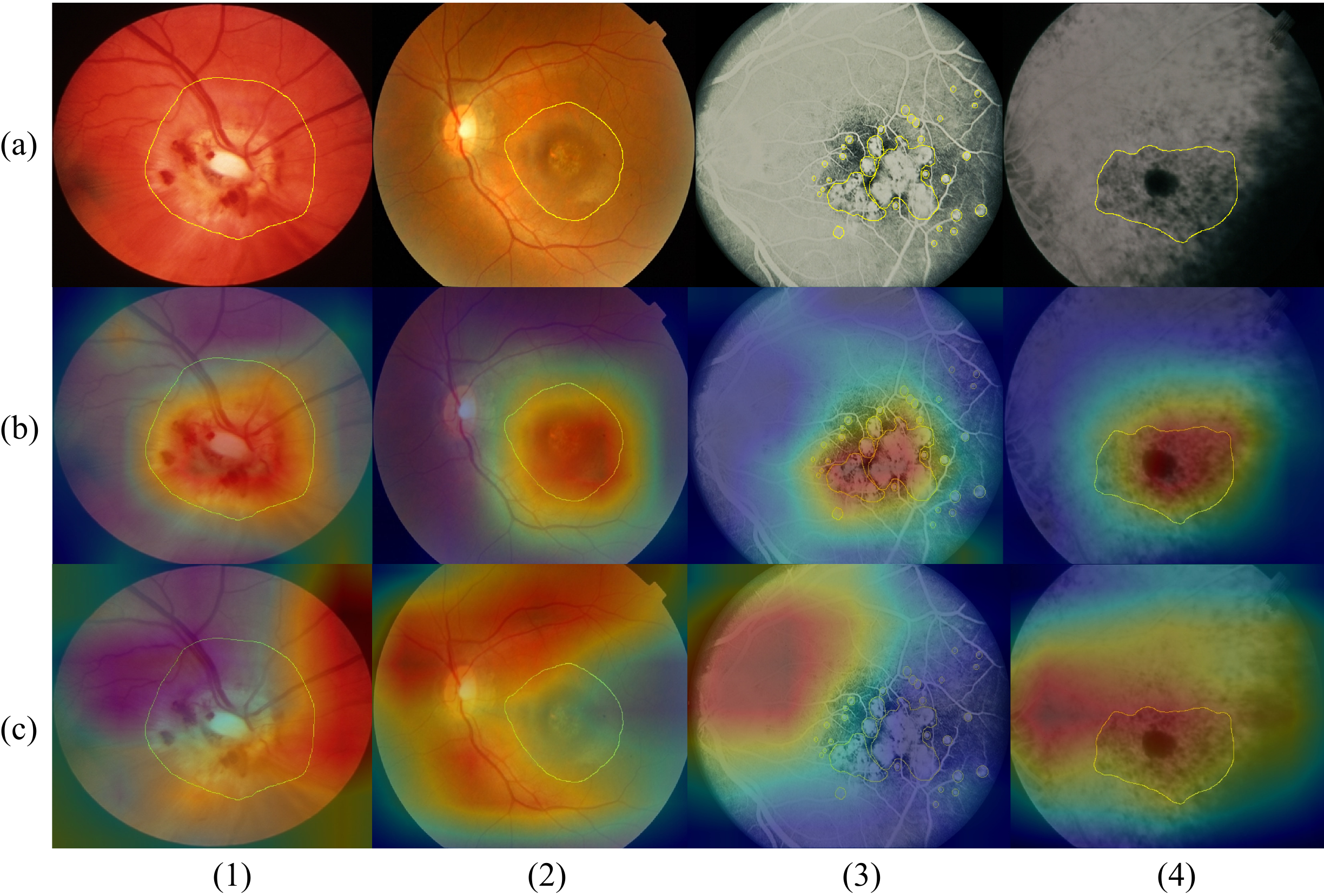}
\end{center}
\vspace{-0.6cm}
    \caption{
    This figure shows the randomly selected qualitative results of CAM. For the detailed explanation, please refer to subsection 5.3.
    }
\label{fig:figure7}
\vspace{-0.2cm}
\end{figure}

The main idea of DNN visual explanation module evaluation is that if our DNN visual explanation results generated by CAM \cite{zhou2015cnnlocalization} are accepted by ophthalmologists, it implies that the proposed method is qualitatively effective. To prove the claim, we build the other retinal image dataset with $300$ retinal images labeled by ophthalmologists and exploit the CNN visualization tool, CAM, to visualize the learned feature and compare it to the ground truth retinal image. We show the qualitative results in Figure \ref{fig:figure7}. In Figure \ref{fig:figure7}, row (a) shows the four different kinds of raw images of retina diseases and each raw image has a yellow sketch labeled by the ophthalmologist to highlight the lesion areas on the retina. The numbers from (1) to (4) denote the four different diseases, including Optic Neuritis, Macular Dystrophy, Albinotic Spots in Macula, and Stargardt Cone-Rod Dystrophy, respectively. We exploit CAM to generate row (b) to demonstrate the visualization results of our DNN-based model. Then, row (c) is produced by the same method as the row (b). Note that both row (b) and row (c) use the same pre-trained weights of ImageNet but row (b) has fine-tuning on DEN dataset and row (c) has no fine-tuning on DEN. The comparison of row (b) and row (c) shows that the DNN-based model successfully learns the robust features of retinal images by training on our DEN dataset. Also, row (b) indicates that the features learned by DNN agree with the domain knowledge of ophthalmologists. That is to say, the activation maps of our deep models are aligned with image features that are clinically recognized by ophthalmologists as linked with the identified disease. The above experimental results show our proposed AI-based method is qualitatively effective.

\section{Conclusion}
To sum up, we propose an AI-based method to automatically generate medical reports for retinal images to improve the traditional retinal diseases treatment procedure. The proposed method is composed of a DNN-based module, including RDI and CDG sub-modules, and DNN visual explanation module. To train our deep models and validate the effectiveness of our RDI and CDG, we propose a large-scale retinal disease image dataset, DEN. Also, we provide another retinal image dataset manually labeled by ophthalmologists to qualitatively evaluate the proposed method. Our experimental results show the proposed AI-based method is effective and successfully improves the conventional treatment procedure of retinal diseases.

\section{Acknowledgments}
This work is supported by competitive research funding from King Abdullah University of Science and
Technology (KAUST) and University of Amsterdam.

{\small
\bibliographystyle{ieee_fullname}
\bibliography{egbib}
}

\end{document}